\documentclass[conference]{IEEEtran}
\usepackage{cite}
\usepackage{amsmath,amssymb,amsfonts}
\usepackage{algorithm}
\usepackage{algpseudocode}
\usepackage{graphicx}
\usepackage{textcomp}
\usepackage{xcolor}
\usepackage{multirow}

\ifCLASSOPTIONcompsoc
\usepackage[caption=false,font=normalsize,labelfont=sf,textfont=sf]{subfig}
\else
\usepackage[caption=false,font=footnotesize]{subfig}
\fi

\def\BibTeX{{\rm B\kern-.05em{\sc i\kern-.025em b}\kern-.08em
    T\kern-.1667em\lower.7ex\hbox{E}\kern-.125emX}}

\begin{document}

\title{I Know How: Combining Prior Policies to Solve New Tasks}

\author{\IEEEauthorblockN{Malio Li*, Elia Piccoli*, Vincenzo Lomonaco, Davide Bacciu}
\IEEEauthorblockA{\textit{Department of Computer Science} \\
\textit{University of Pisa}\\
\texttt{malio.li@di.unipi.it, elia.piccoli@phd.unipi.it}}
}


\maketitle

\IEEEoverridecommandlockouts
\IEEEpubid{\makebox[\columnwidth]{ 979-8-3503-5067-8/24/\$31.00~\copyright2024 IEEE \hfill} 
\hspace{\columnsep}\makebox[\columnwidth]{ }}
\IEEEpubidadjcol

\begin{abstract}
Multi-Task Reinforcement Learning aims at developing agents that are able to continually evolve and adapt to new scenarios.
However, this goal is challenging to achieve due to the phenomenon of catastrophic forgetting and the high demand of computational resources.
Learning from scratch for each new task is not a viable or sustainable option, and thus agents should be able to collect and exploit prior knowledge while facing new problems.
While several methodologies have attempted to address the problem from different perspectives, they lack a common structure.
In this work, we propose a new framework, I Know How (IKH), which provides a common formalization.
Our methodology focuses on modularity and compositionality of knowledge in order to achieve and enhance agent’s ability to learn and adapt efficiently to dynamic environments.
To support our framework definition, we present a simple application of it in a simulated driving environment and compare its performance with that of state-of-the-art approaches.
\end{abstract}


\IEEEpeerreviewmaketitle

\section{Introduction}
Reinforcement Learning (RL) has achieved significant milestones across several domains, especially with a notable focus and effort on games.
RL-trained agents have shown outstanding performance, often surpassing human-level, in various games with different degrees of complexity.
A critical limitation is that these agents are outstanding in solving a specific task but cannot adapt and transfer their abilities to new scenarios.
Moreover, the resource-intensive nature of training these models further increases the challenge, obstructing their scalability and applicability in dynamic environments.

Real-world problems constantly evolve and redefine their requirements and domains. The ability of agents to adapt to continually evolving scenarios becomes crucial.
In the gaming industry, titles undergo iterative development cycles where they introduce new features and levels.
The possibility of designing AI bots that are able to progress concurrently with game development would be a huge benefit for companies in terms of time and resources.
In robotics, agents are required to solve several problems with increasing complexity while retaining the capacity to complete simple tasks.
In these scenarios agents should be able to re-use prior knowledge, acquired by learning simple problems, to solve new tasks and adapt to changes of the environment possibly reducing the cost needed to re-train the models from scratch.

\begin{figure}[!h]
    \centering
    \includegraphics[scale=0.26]{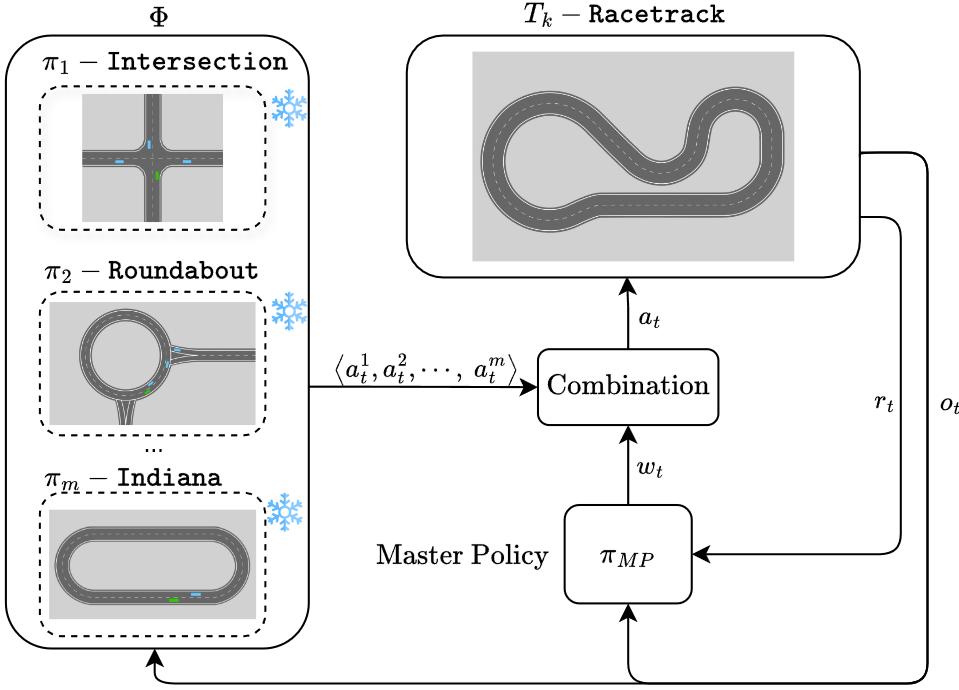}
    \caption{Illustration of the implementation of the IKH framework used in this work. Given a set $\Phi$ of $m$ pre-trained policies on auxiliary tasks, $\pi_{MP}$ predicts the weights $w$ to assign at each action from the policies to define agents' behavior.}
    \label{fig:weighted action}
\end{figure}

To address these challenges, we propose a novel approach called \textit{I Know How (IKH)} (\figurename 
 \ref{fig:weighted action}).
Our approach leverages a repository of pre-trained policies and reasons in the policies' action space to combine their behaviors to solve a novel task.
The IKH framework offers scalability, facilitating the integration of new policies and adaptation to new scenarios. Furthermore, IKH ensures by design explainability on the decision-making processes of the agent.

This paper presents the ideas underlying the proposed approach, its implementation, and an initial empirical evaluation and comparative analysis with state-of-the-art methods.

\section{Related works}
Multi-task learning has been widely studied in several RL scenarios addressing the problem via different perspectives.
Progressive Neural Networks (PNN) \cite{pnn} learn sequentially different tasks instantiating for each one a new column. The $n^{th}$ column receives lateral connections from all the previous columns. While this approach helps to avoid catastrophic forgetting, it leads to more than linear scaling complexity and requirement of resources. Other works \cite{progcomp, modnet} try to reduce the amount of resources needed by distilling and compressing the new column in a knowledge base or using task-driven priors to limit the combination space between different tasks.

A different set of solutions enhances agents with prior knowledge following the \textit{option} \cite{option} framework definition, skills as policies, that can be learned and used in several ways.
They can be learned independently and then composed \cite{l2c} or leveraged to distill a central policy that must resemble the different skills \cite{distral}.
Other methods \cite{spirl, simpl, sbmbrl} learn latent representations, which represent the skills’ embeddings or dynamics, that are later exploited to plan in the skill space or to easily adapt to new tasks via similarity between representations.

Hierarchical RL approaches aim to address complex decision-making tasks by decomposing them into hierarchical subtasks, enabling agents to learn and navigate efficiently through multiple levels of abstraction.
H-DRLN \cite{hdrln} defines a set of policies that are combined with simple actions adding the possibility to alternate between complex and simple behaviors.
ULTRA \cite{ultra} via Unsupervised RL learn a set of transferable sub-policies that are used by a meta-learner to adapt to new scenarios, separating the model in task agnostic and task-specific parts.
In \cite{snn} authors pre-train subpolicies via Stochastic Neural Network and for each new task a new manager is trained to dictate which policy to use or to condition for next $\mathcal{T}$ steps. 
Ensemble methods \cite{ensemble_ddpg, deep_ensemble} usually are used to accelerate and stabilize agents' learning process averaging the output of $k$ policies trained on the same task.

This work collects the main take-aways from these methods proposing a novel framework to incorporate prior policies while maintaining scalability and explainability.



\section{I Know How (IKW)}
Multi-task learning scenarios focus on agents that are required to solve a sequence of tasks.
Humans, when faced with a similar challenge, often rely on their abilities to solve similar problems, thus drawing from past experiences and knowledge.
Our approach is inspired by this behavior in order to achieve agents that can learn and adapt in ways that are more akin to human learning.

The key components that define our \textit{IKH framework} are:
\begin{enumerate}
    \item A set of \textit{pre-trained policies}, $\Phi = \langle \pi_1, \pi_2, \dots, \pi_m \rangle$.
    The agent has access to $\Phi$ that is composed by $m$ pre-trained policies, which have been previously trained.
    \item \textit{Master Policy (MP)}, $\pi_{MP}^\theta$.
    This model with parameters $\theta$ will be learned during the training process, its objective is to control how pre-trained policies are combined.
    \item \textit{Combination} module, $\mathcal{C}$.
    Represents a function that combines the output of pre-trained models and master policy to compute the actual agent's action.
\end{enumerate}

While the policies in $\Phi$ are specialized to solve specific problems, with possibly different objectives, $\pi_{MP}^\theta$ learns how to leverage the pre-trained modules based following the dynamics defined by $\mathcal{C}$. 
Differently from the standard definition of policy, $\pi_{MP}^\theta$ does not output the final action but instead, the information needed to combine the outputs from the pre-trained policies.
Thus, exploiting the interaction with the environment and the optimization process MP eventually learns how to optimally exploit the prior knowledge to solve the new task.

\begin{algorithm}[!t]
    \caption{\textit{IKH} training loop}\label{master policy training loop}
    \begin{algorithmic}[1]
        \Require Environment $\mathcal{E}$, Pre-trained Policies $\Phi$, Master Policy $\pi_{MP}^\theta$, Combination module $\mathcal{C}$, RL algorithm \textit{Alg}
        \State Load and freeze $\Phi$
        \For {$N$ steps}
            \State Observe the current state $s_t$ from $\mathcal{E}$
            \State ${o_{MP}}_t = \pi_{MP}^\theta(s_t)$ \label{mp_out}
            \State $\langle a_t^1, a_t^2, \dots, a_t^m \rangle = \Phi (s_t)$ \label{skill_out}
            \State $a_t = \mathcal{C}(\langle a_t^1, a_t^2, \dots, a_t^m \rangle, {o_{MP}}_t)$ \label{comb}
            \State make a step in $\mathcal{E}$ with action $a_t$
            \State observe $s_{t+1}$ and get reward $r_t$
            \State store the transition $\langle s_t, {o_{MP}}_t, r_t, s_{t+1} \rangle$
            \If{\textit{update condition}}
                \State update $\pi_{MP}^\theta$ according to \textit{Alg}.
            \EndIf
            \If{\textit{episode completed}}
                \State reset $\mathcal{E}$.
            \EndIf
        \EndFor
    \end{algorithmic}
\end{algorithm}

\textbf{Algorithm \ref{master policy training loop}} shows the general training loop for our framework.
It first loads the set of pre-trained policies, then it performs $N$ interaction with the environment.
At each step, it computes the outputs of the models based on the current state - lines \ref{mp_out}, \ref{skill_out} - and in line \ref{comb} determine the actual action.
The transition tuples are then stored to later optimize the master policy based on the RL algorithm of choice.
It is worth mentioning that this approach is agnostic with respect to the RL algorithm employed during the training process.

The structure of this framework allows its main components to be implemented with different strategies while maintaining its original definition.
For example, $\Phi$ can be defined by incrementally adding more policies as the agent faces new tasks, or they can be learned via offline learning using collected data, etc.
The \textit{master policy} can be defined to reason only on the current state of the environment or to leverage information - i.e. embeddings - from the pre-trained policies to condition its prediction.
The \textit{combination} module may be described by a function or parameterized as a neural network and learned during the training loop.
While these different definitions does not interfere with the framework structure, they may be crucial in determining agents' capabilities.\\

In this work, we consider a simple and initial setting of the \textit{IKH} framework (\figurename \ref{fig:weighted action}).
In particular, we assume a variable set of pre-trained policies specialized on simple scenarios.
The master policy learns a non-linear mapping from the state space to a latent space of the same size as $|\Phi|$, representing a set of weights $w$.
These weights are then used to compute the agent's action as a linear combination of the pre-trained policies' actions according to the following formula:
\begin{equation}\label{final action}
    a_t = \cfrac{\sum_{i=1}^m a_t^i w_t^i}{W_t}
\end{equation}
where $W_t = \sum_{i=0}^m w_t^i$, $m = |\Phi|$.
In this scenario we are allowed to compute the action as a weighted average since the pre-trained policies' actions are also valid for the new task, they share the same state and action space.
Generally, this assumption might not hold so more complex strategies must be used to adapt the behaviors to be coherent with the MDPs on which they were originally defined and the current one.

\section{Experiments}\label{sec: experiments}

\begin{figure}[!t]
    \centering
    \includegraphics[width=0.25\textwidth]{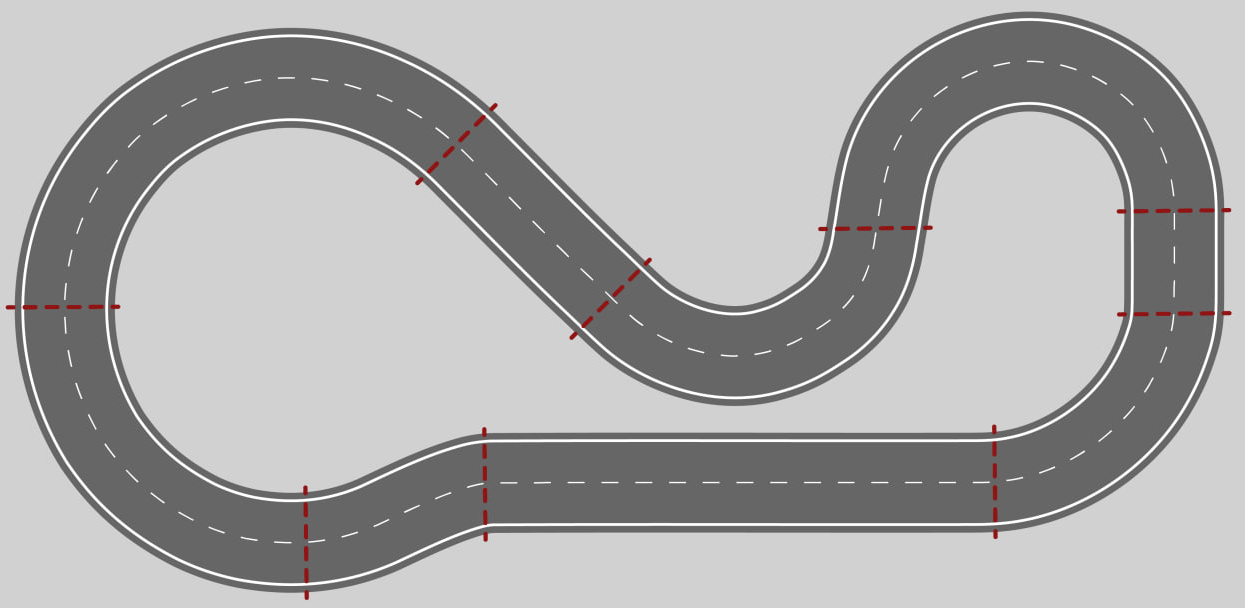}
    \caption{The \texttt{Racetrack} scenario used to evaluate agents' performance. The track is divided into nine sectors, delimited by the red lines, based on the different characteristics of the circuit.}
    \label{fig: racetrack}
\end{figure}

\subsection{Experimental Setup}\label{setup}
We used \texttt{stable-baselines3} (SB3) \cite{sb3} as a backbone, and we built our framework\footnote{https://github.com/xiaoli98/I-Know-How} as an extension built on top of SB3's APIs.
To evaluate the performance of the proposed approach, we used \texttt{Highway-Env}\cite{highwayenv}
environment, which offers different road scenarios and it is easily customizable. 
The observation space is a kinematics observation with a dimension of 45 and the action space is 2.
The list of road scenarios used to train basic skills is: \texttt{Highway} (H), \texttt{Merge} (M), \texttt{Roundabout} (R), \texttt{Intersection} (I), \texttt{U-turn} (U). 
We additionally build two other custom road scenarios called \texttt{Indiana} (IN) and \texttt{Lane centering} (L).
The first resembles the Indianapolis speedway, while the second one is a sinusoidal road changing the frequency and amplitude at each episode.
Some of the environments are reported in \figurename \ref{fig:weighted action}, and all the standard ones can be viewed on the documentation page.
We populate $\Phi$ pre-training policies (Appendix \ref{appendix: pre-trained policies}) to solve these simple environments, using SAC.
The master policy $\pi_{MP}^\theta$ is implemented using a simple MLP with 2 layers of 256 units and ReLu as activation function.
To train and evaluate the performance of our methodology, instead, we use the \texttt{Racetrack} scenario which is a complex scenario composed of several corners with different degrees and straight parts (\figurename \ref{fig: racetrack}).
In order to compare the performance of our approach we tested two state-of-the-art algorithms:
\begin{enumerate}
    \item A single agent trained with SAC to solve \textit{only} the objective task from scratch.
    \item A multi-task agent leveraging the PNN structure.
    The frozen columns correspond to the pre-trained policies from $\Phi$.
    The last column is trained to solve the objective task while leveraging the adapters from previous columns.
    This setting is more similar to ours, benefiting from knowledge transfer from previous columns.
\end{enumerate}
Total number of steps and gradient updates is the same (10 million steps and 10 gradient steps for each update).
The choice of using total steps and not total episodes is due to the difference in episode length between models.

\subsection{Results}
To evaluate agents' performance, the reward is not the best metric.
In this scenario, the reward function provided by the environment is composed of different elements, i.e. cost per action, line-centering incentive, collision, etc.
Just looking at the reward we might get a high reward with sub-optimal behaviors with respect to the final objective of being able to complete the \texttt{Racetrack}.
To quantitatively compare the performance of different methods the metric used is the average number of sectors completed by the vehicle.
In \figurename\ref{fig: racetrack} in red are delimited the 9 different sectors.
At each episode, both during training and test, the agent is randomly spawned in one of these sections with a random number of vehicles inside the track (up to 5).
Hitting another car or going outside the track results in the termination of the episode.
To collect results we average the performance over 100 episodes for 5 different seeds.

\begin{figure}[!t]
    \centering
    \includegraphics[width=0.3\textwidth]{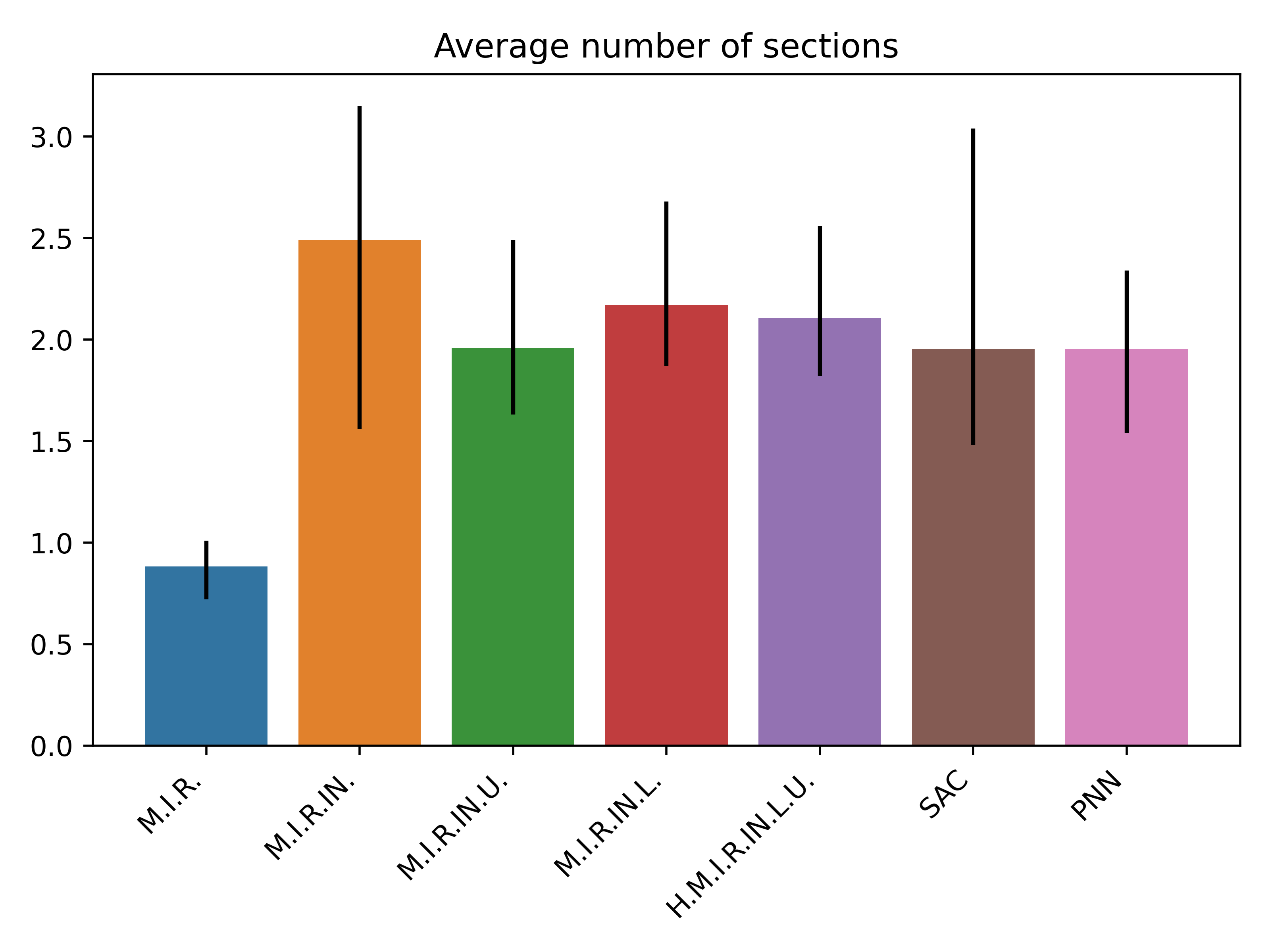}
    \caption{Average number and variance of sections of different agents using varying number of pre-trained policies compared to baseline methods.
    The acronym reported below refers to the name of the environments (see \ref{setup}) that are solved by the policies inside $\Phi$.}
    \label{fig: ablation skills}
\end{figure}

\figurename \ref{fig: ablation skills} reports the average performance for different combinations of pre-trained policies while comparing to baseline methods.
The acronym refers to the tasks that are solved by the policies set.
While different combinations with different sizes of $\Phi$ on average perform similarly to SAC or PNN, \texttt{M.I.R.IN.} performs best.
This showcases that not necessarily a larger number of pre-trained policies leads to better performance.

The average number of sections is relatively low with respect to the total length of the track.
To better understand agents' behavior we look at the distribution of completed sectors across the evaluation phase.
The mean and variance of each completed sectors are reported in Table \ref{tab: passed sectors statistics} (more details are presented in Appendix \ref{appendix: boxplot}).
While baseline approaches struggle to solve the complete track, our method on average can solve it more consistently.
The distribution shift towards a small number of sectors may be related to different factors such as: \textit{(i)} the number of cars spawned in the track that obstruct agents' movement, \textit{(ii)} poor generalization of some section given by the randomness in the training process or \textit{(iii)} challenging and more complex scenarios, i.e. tight corners.

\begin{table}[!bt]
\centering
\caption{Distribution of Number of Sectors Completed}
\label{tab: passed sectors statistics}
\begin{tabular}{|c|ccc|}
\hline
\multirow{2}{*}{\textit{\textbf{\# Completed Sectors}}} & \multicolumn{3}{c|}{\textit{Methods}}                                                                              \\ \cline{2-4} 
                                 & \multicolumn{1}{c|}{IKH}                        & \multicolumn{1}{c|}{SAC}              & PNN              \\ \hline
0-2                              & \multicolumn{1}{c|}{60.65 $\pm$ 8.90}          & \multicolumn{1}{c|}{61.85 $\pm$ 9.10} & 60.00 $\pm$ 7.30 \\ 
3-5                              & \multicolumn{1}{c|}{25.37 $\pm$ 5.29}          & \multicolumn{1}{c|}{33.15 $\pm$ 6.37} & 36.25 $\pm$ 6.59 \\ 
6-8                        & \multicolumn{1}{c|}{\textbf{9.32 $\pm$ 4.22}} & \multicolumn{1}{c|}{4.77 $\pm$ 6.38}  & 3.75 $\pm$ 3.25  \\
9 or more & \multicolumn{1}{c|}{\textbf{4.65 $\pm$ 2.91}} & \multicolumn{1}{c|}{0.22 $\pm$ 0.83} & 0.00 $\pm$ 0.00\\ \hline
\end{tabular}
\end{table}

\subsection{Explainability}
An additional feature of the framework given by its structure, in particular by the combination module, is the explainability.
In fact, it is possible to analyze how the different pre-trained policies are exploited by agents to solve the task.

\begin{figure}[!t]
\centering
\subfloat[\label{fig:frame1}]{
    \includegraphics[width=0.13\textwidth]{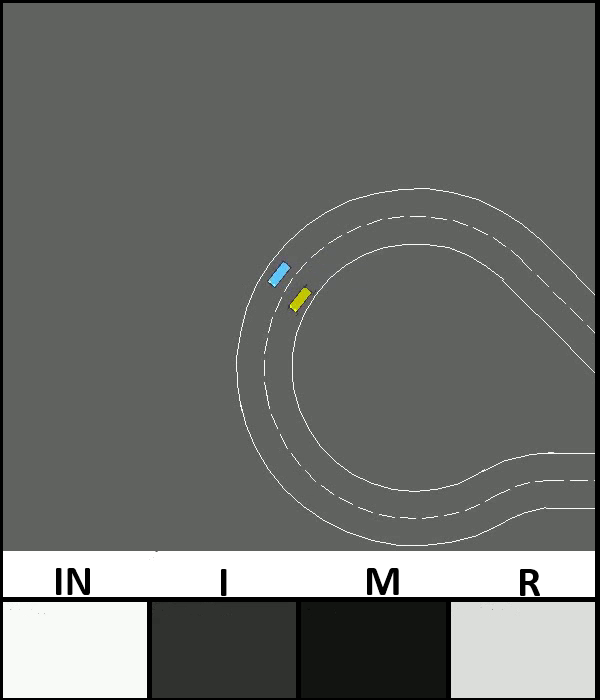}   
}
\hfill
\subfloat[\label{fig:frame2}]{
    \includegraphics[width=0.13\textwidth]{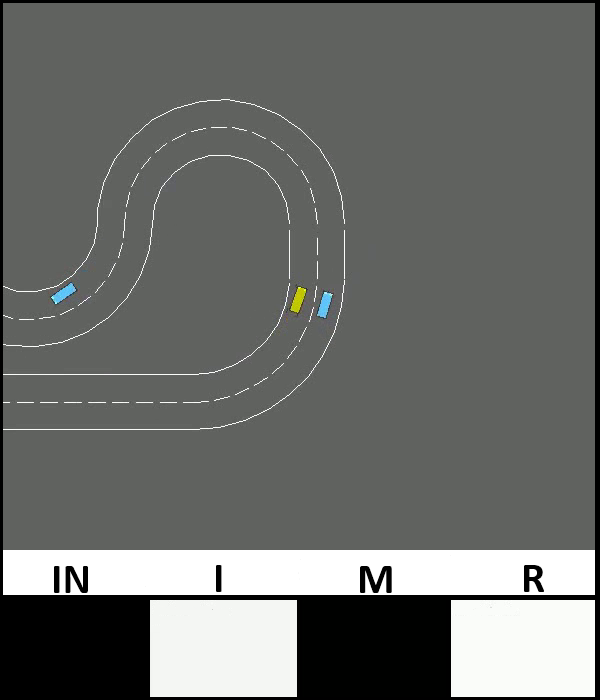}
}
\hfill
\subfloat[\label{fig:frame3}]{
    \includegraphics[width=0.13\textwidth]{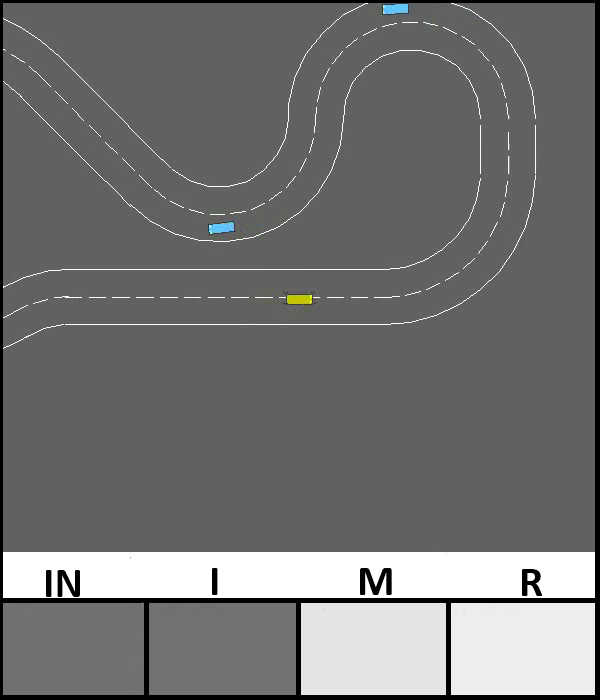}
    
}
\caption{Each image portrays the agent (yellow car) in different part of the track. Each weight predicted by $\pi_{MP}$ is bounded between $[0,1]$. Mapping the upper bound to \textit{white} and lower bound to \textit{black} it possible to appreciate how different pre-trained policies are exploited in different scenarios.}
\label{fig: frames}
\end{figure}

In \figurename\ref{fig: frames} we report three frames of the \texttt{M.I.R.IN.} configuration in different parts of the track.
As additional information below is reported the four different pre-trained policies used in this setting, where they are colored based on the weight assigned by the master policy.
During the wide corner \figurename\ref{fig:frame1}, the policies for \texttt{R} and \texttt{IN} are highly activated.
In \figurename\ref{fig:frame2} at the end of the corner \texttt{I} and \texttt{R} are activated since both can handle changes between corners and straights.
In the last frame \figurename\ref{fig:frame3} during a straight road scenario all policies contribute, this part of the track is easier and all of them can manage this scenario.


\section{Conclusions}
This paper presents a novel framework to address the complexities of multi-task RL problems by exploiting prior knowledge.
Our framework is structured around three fundamental components: a set of \textit{pre-trained policies}, a \textit{master policy} to link observations and policies, and a \textit{combination} module that integrates information from former components.

We present a practical application of our framework in a simulated driving environment, presenting the implementation of different components and showcasing its adaptability.
Moreover, we conducted comparisons with state-of-the-art methods, where our approach achieves competitive results while offering distinct advantages.
By leveraging pre-trained policies and optimizing the latent relationship between observations and policies, our framework exhibits promising performance highlighting its potential for real-world applications.

Looking forward, first, we would like to explore more complex and challenging scenarios to test the limitations of our approach.
Additionally, our framework opens different research paths and challenges based on its underlying mechanisms.
For example, how to dynamically extend the pool of pre-trained policies following some initial promising directions \cite{polsubspace}.
This process may involve exploring alternative architectures and methods, and more importantly, incorporating insights from related fields such as transfer learning and meta-learning.

We recognize the importance of studying the challenges to employ our framework in real-world settings.
This entails investigating issues such as efficiency, environmental variability robustness, and large-scale task scalability.
We believe this is a promising research direction that paves the way for future research works from which also game development and AI bot design may benefit.

\section*{Acknowledgment}
Research partly funded by PNRR-  PE00000013 - "FAIR - Future Artificial Intelligence Research" - Spoke 1 "Human-centered AI", funded by the European Commission under the NextGeneration EU programme, and by the Horizon EU project EMERGE (grant n. 101070918).

\appendices
\section{Pre-trained policies}\label{appendix: pre-trained policies}
\begin{figure*}[!tbh]
\centering
\subfloat[\label{fig: highway_rewMean}]{
    \includegraphics[width=0.4\textwidth]{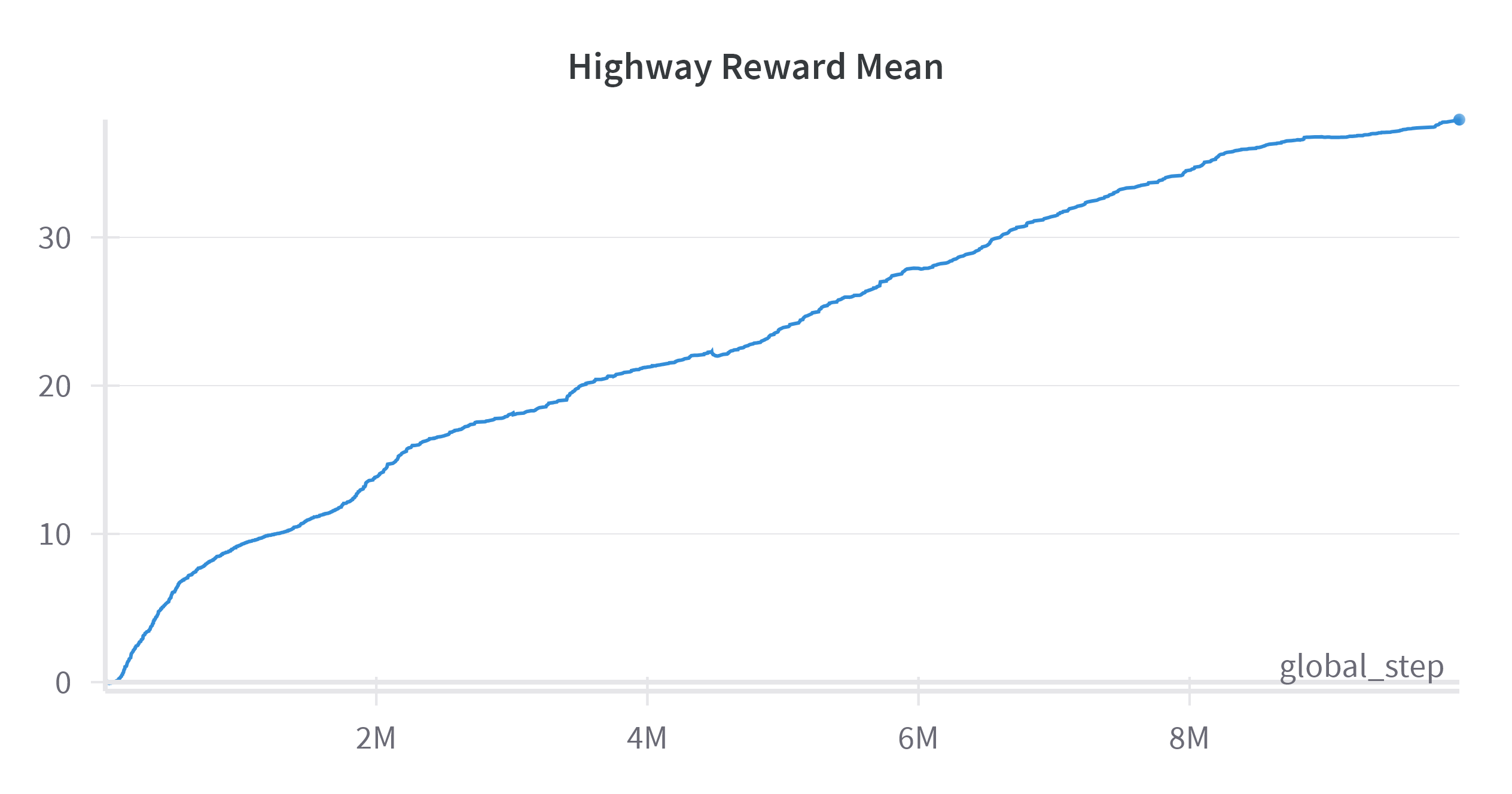}
}
\hfill
\subfloat[\label{fig: merge_rewMean}]{
    \includegraphics[width=0.4\textwidth]{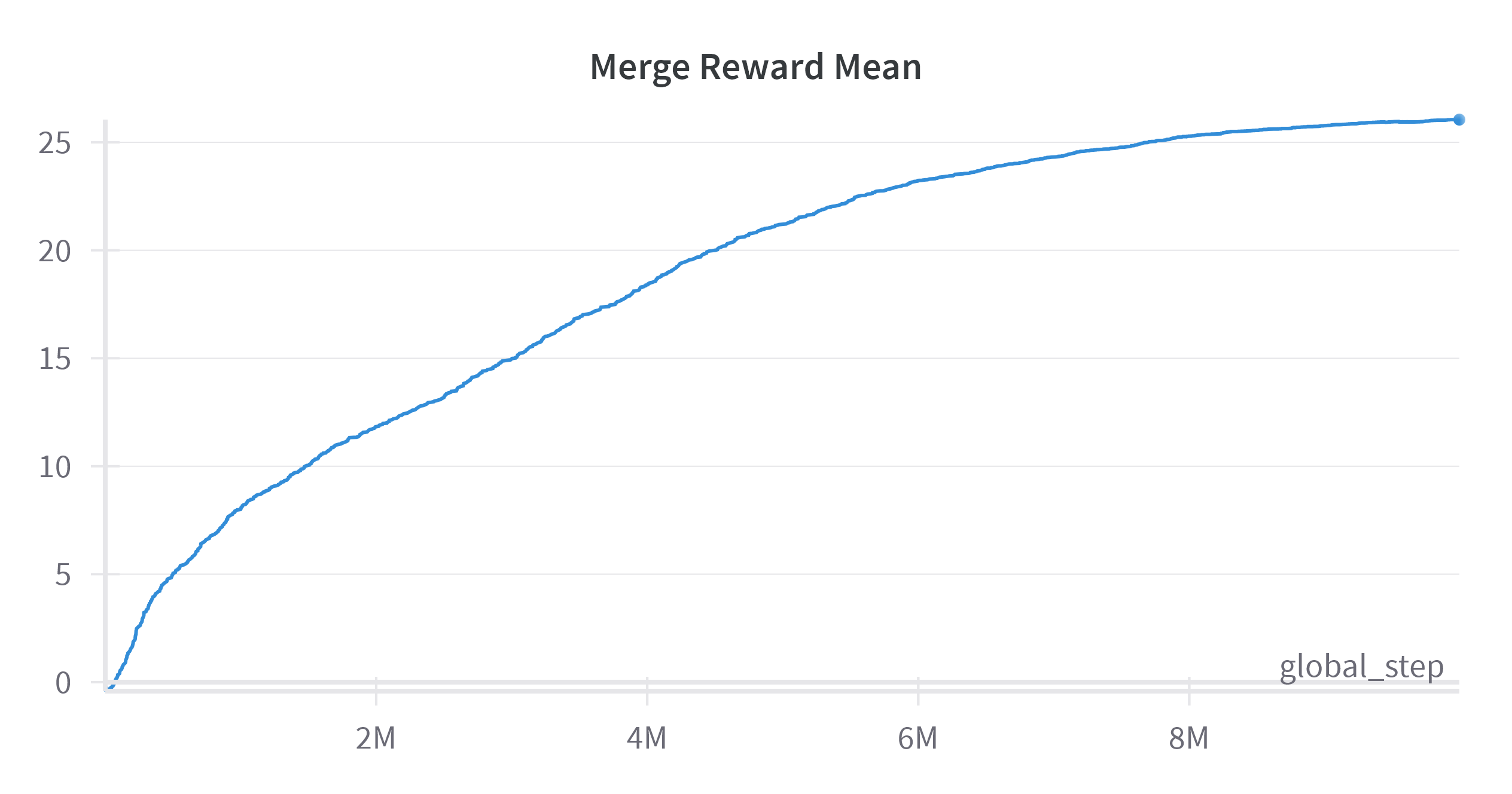}
}
\hfill
\subfloat[\label{fig: indiana_rewMean}]{
    \includegraphics[width=0.4\textwidth]{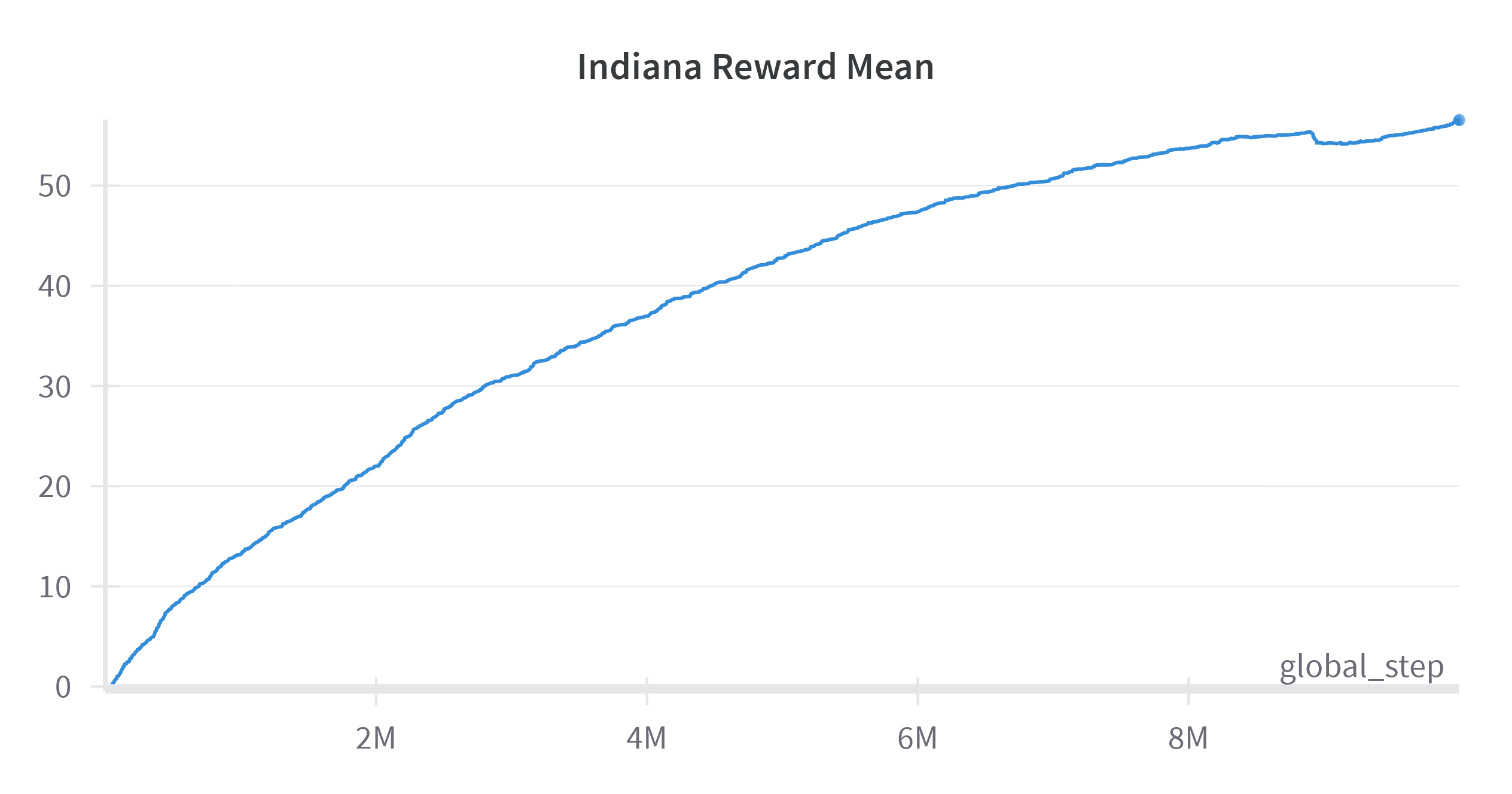}
}
    \hfill
    \subfloat[\label{fig: intersection_rewMean}]{
        \includegraphics[width=0.4\textwidth]{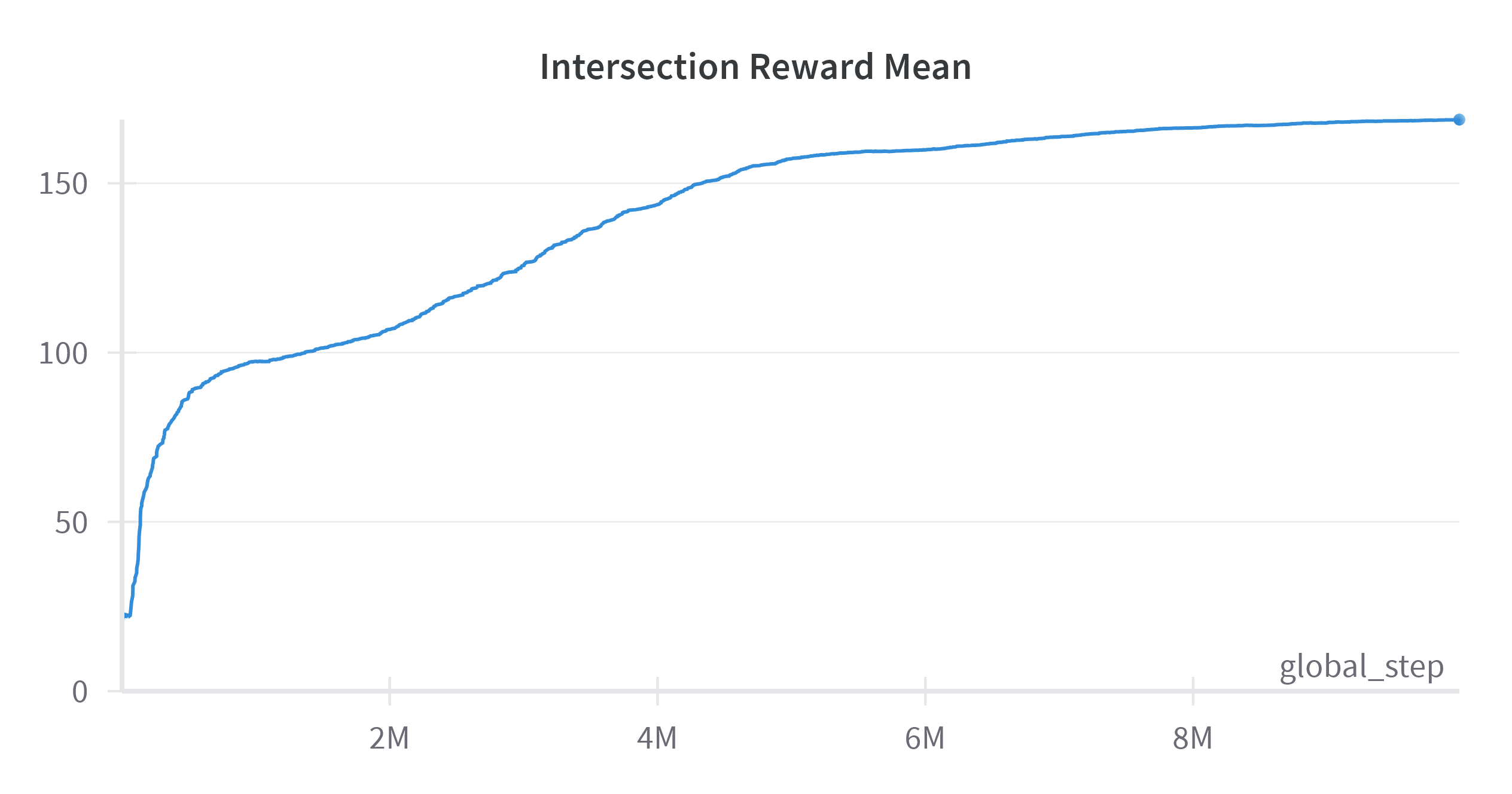}
    }
    \hfill
    \subfloat[\label{fig: lane_centering_rewMean}]{
        \includegraphics[width=0.4\textwidth]{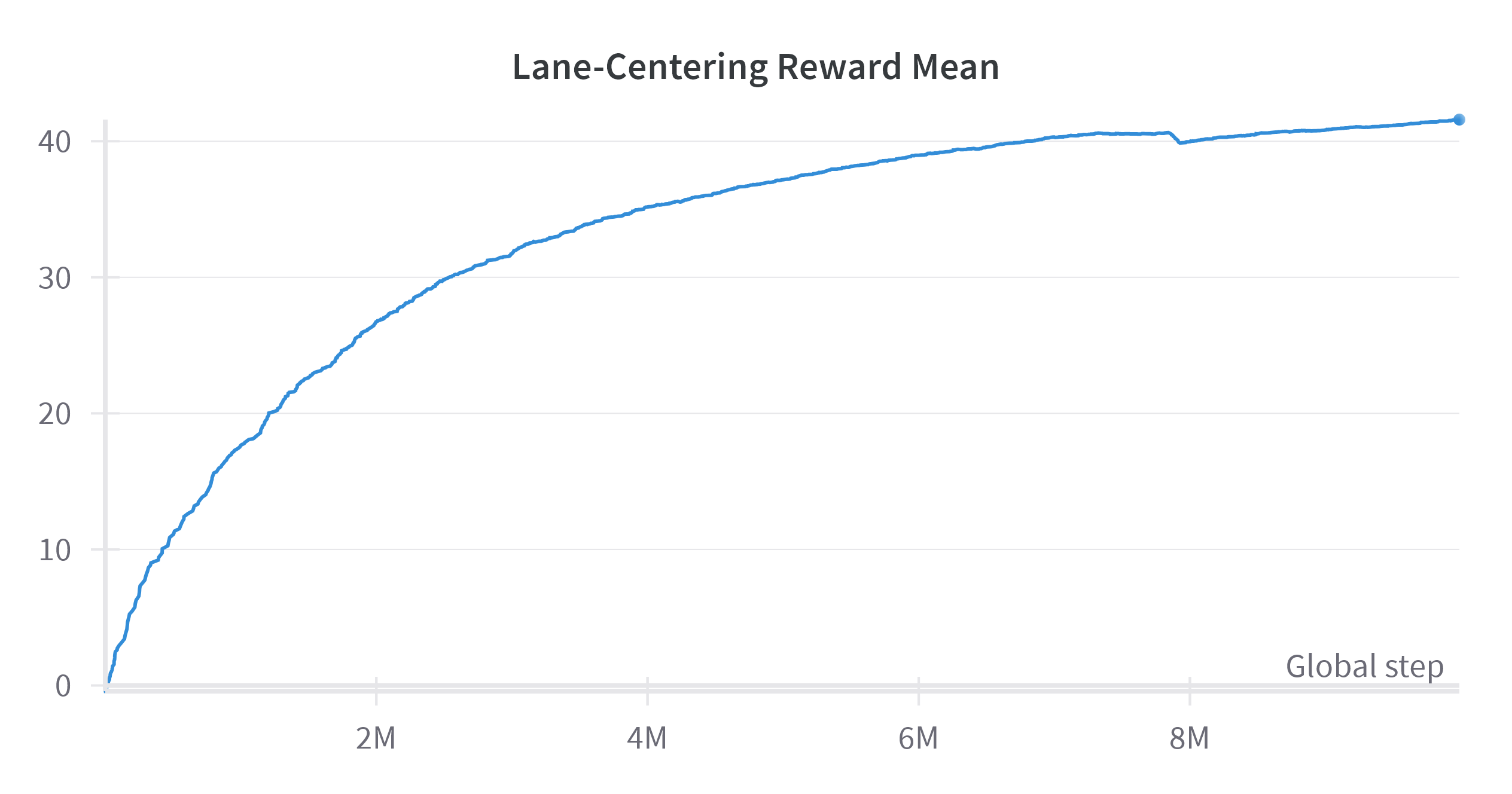}
    }
    \hfill
    \subfloat[\label{fig: roundabout_rewMean}]{
        \includegraphics[width=0.4\textwidth]{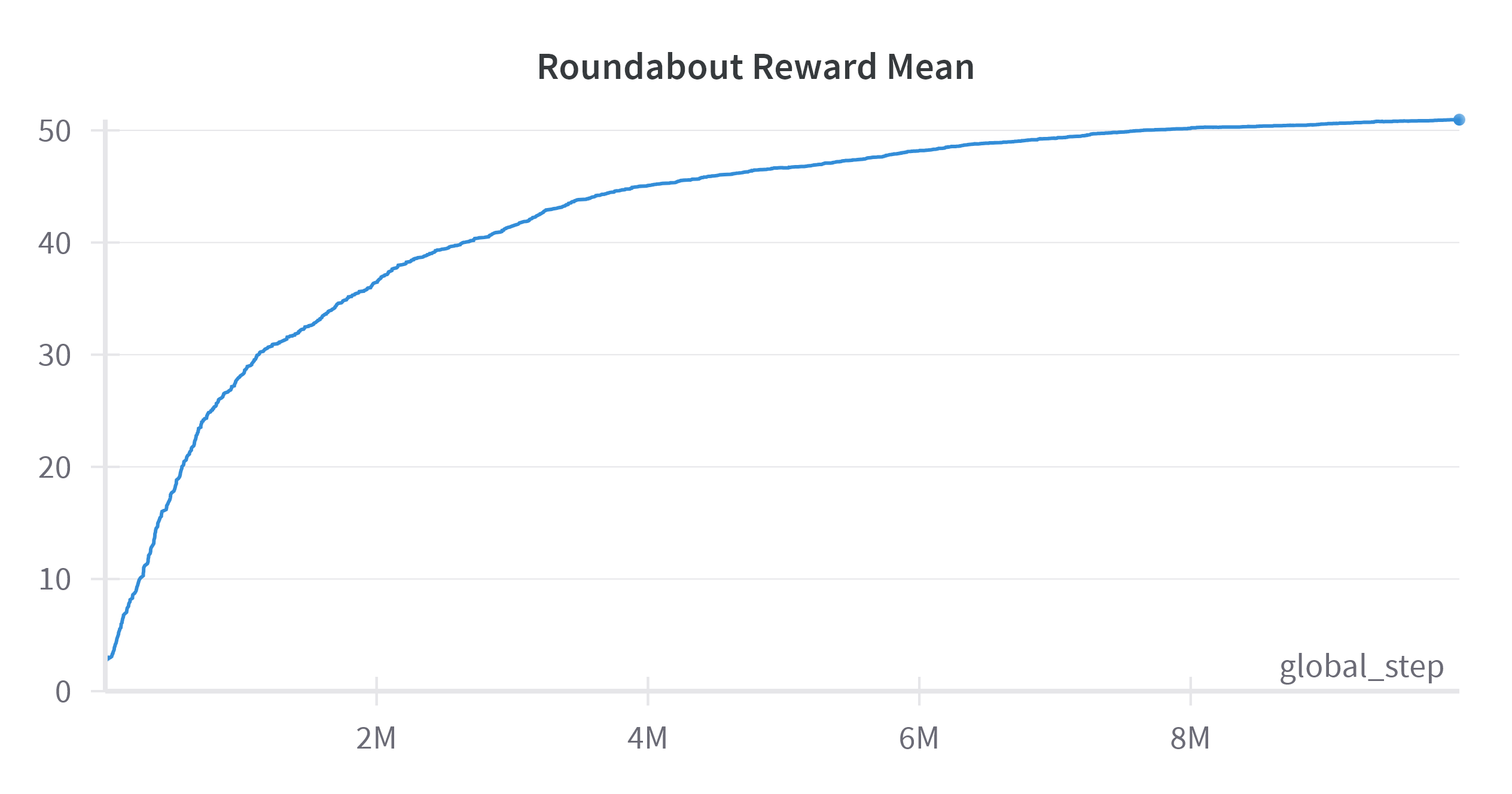}
    }
    \hfill
    \subfloat[\label{fig: uturn_rewMean}]{
        \includegraphics[width=0.4\textwidth]{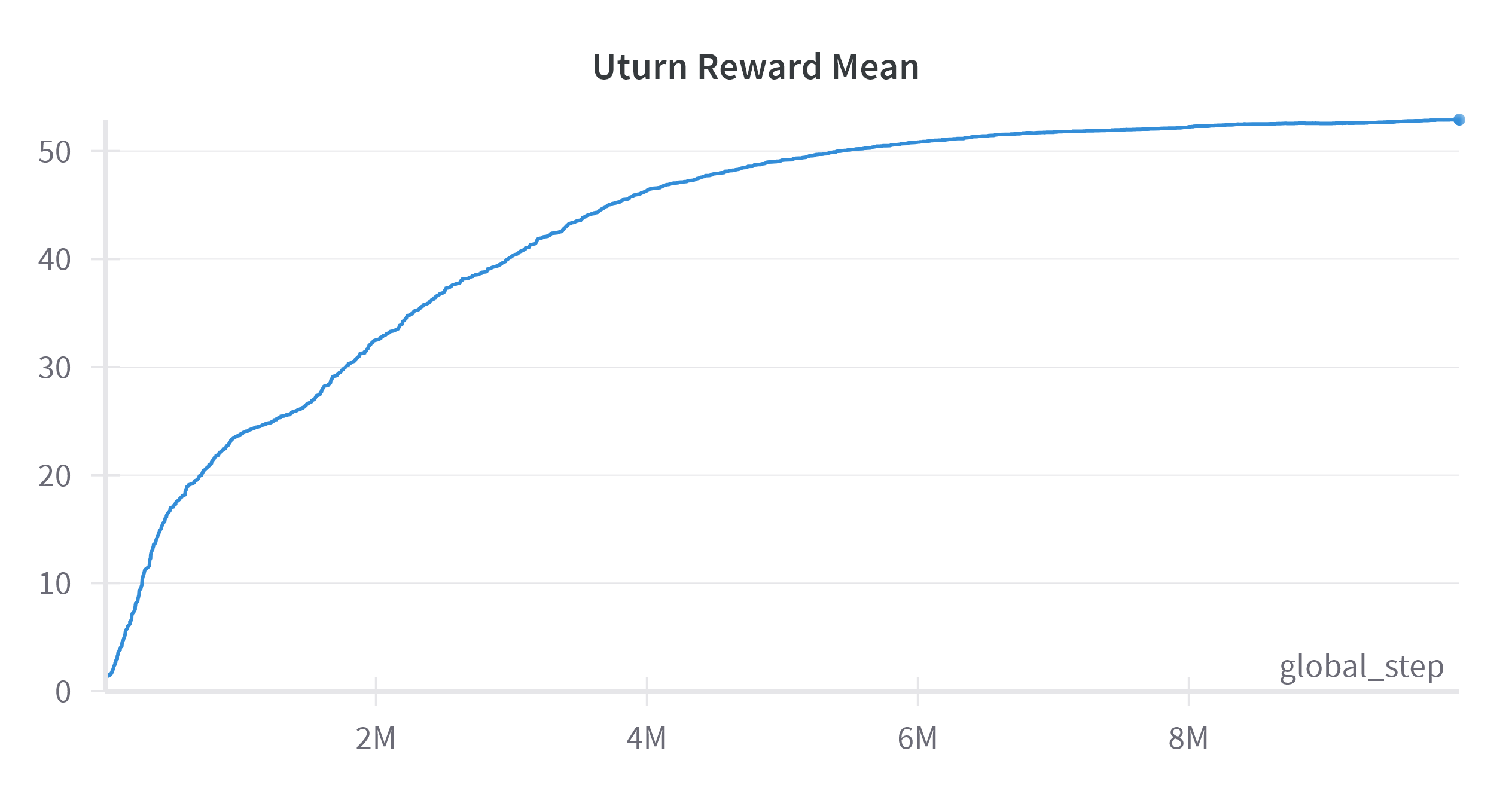}
    }
\caption{Reward graphs of pre-trained policies.}
\label{fig: subpolicies reward}
\end{figure*}

Here we provide additional details on the training of the set of pre-trained policies. The set of pre-trained policies are trained using the Soft Actor-Critic algorithm.
Table \ref{hyper parameter list} shows the hyperparameter configuration of pre-trained policies, from top to bottom learning rate (lr), batch size (bs), gradient steps (gs), total steps (ts), soft update coefficient ($\tau$), discount factor ($\gamma$) and entropy coefficient ($\alpha$). 
\figurename \ref{fig: subpolicies reward} shows the episodic reward graphs during the training process, respectively: \texttt{Highway}, \texttt{Merge}, \texttt{Indiana}, \texttt{Intersection}, \texttt{Lane-centering}, \texttt{Roundabout} and \texttt{U-turn}. 
To speed up the training, we used 16 parallelized environments per training, which reduces the time to collect transitions using SAC.

The characteristics of the machine used to perform the training are listed below:
\begin{itemize}
    \item Operating System: Ubuntu 20.04.4 LTS x86 64bit  
    \item 56 CPUs: Intel(R) Xeon(R) Gold 6238R @ 2.20Ghz
    \item RAM: 448 GiB
    \item GPU: Nvidia Tesla T4 16GiB
\end{itemize}

\section{Additional environment description}
We have mentioned in Section \ref{sec: experiments} that the observation space has dimension 45. This is obtained from flattening a matrix of dimensions $5\times9$, where 5 refers to observed vehicles (including the controlled one), and 9 are the monitored features:
\begin{itemize}
    \item 1 value for presence, a bit representing if a vehicle is present, it is always 1 for the first row (the controlled vehicle).
    \item 2 values for the $(x,y)$ coordinates of the vehicle.
    \item 2 values for the $(vx, vy)$
 velocity of the vehicle on the axis $x$ and $y$.
    \item 1 value for heading of the vehicle, expressed in radians.
    \item 3 values for longitudinal, lateral, and angular offset of the vehicle on the closest lane
\end{itemize}

The environment reset occurs when the controlled vehicle goes out of track, collides with another vehicle, or when the maximum episode length is reached.

The reward function considers mainly 2 factors: the vehicle's speed and how far the vehicle is from the lane center. 
The reward received at each step is normalized. 
Since the episode length is different for different environments, the total reward changes.
For \texttt{Intersection}, by design, the main objective is to avoid collision by cooperating with other agents and the reward is a cumulative value of several agents, thus the total reward is much higher, resulting in a shift of the total reward of agents.
When the vehicle goes off track a negative reward is given to penalize, and in \texttt{Roundabout} the penalization is heavier.

The frequency at which agents interact with the environment and take an action is 5 Hz.

\begin{table}[!hbt]
\centering
\caption{hyper-parameters of agents trained on \texttt{Highway} (H), \texttt{Merge}(M), \texttt{Intersection} (I), \texttt{Indiana} (IN), \texttt{Lane-centering} (L), \texttt{Roundabout} (R), \texttt{U-turn} (U) and \texttt{Racetrack} (RT) }
\label{hyper parameter list}
\begin{tabular}{|l|l|l|l|l|l|l|l|l|} 
\hline
 &  H & M & I & IN & LC & R & U & RT \\ 
\hline
lr &  1e-4& 1e-4 & 1e-4 & 5e-5 & 1e-4 & 1e-4 & 5e-5 & 1e-4 \\ 
\hline
bs & 1024 & 1024 & 1024 & 8192 & 1024 & 1024 & 4096 & 1024 \\ 
\hline
gs & 10 & 10 & 10 & 10 & 10 & 10 & 10 & 10 \\ 
\hline
ts & 1e7 & 1e7 & 1e7 & 1e7 &
 1e7 & 1e7 & 1e7 & 1e7 \\
\hline
$\tau$ &  0.9 & 0.005 & 0.9 & 0.1 & 0.005 & 0.9 & 0.9 & 0.9\\ 
\hline
$\gamma$ & 0.99 & 0.99 & 0.99 & 0.65 & 0.99 & 0.99 & 0.65 & 0.99\\ 
\hline
$\alpha$ & 0.5 & 0.5 & 0.5 & 0.5 & 1 & 0.5 & 0.5 & 0.5 \\ 
\hline
\end{tabular}
\end{table}

\section{On number of passed sectors}\label{appendix: boxplot}

\begin{figure*}[!tbh]
    \centering
    \subfloat[]{
        \includegraphics[width=0.4\textwidth]{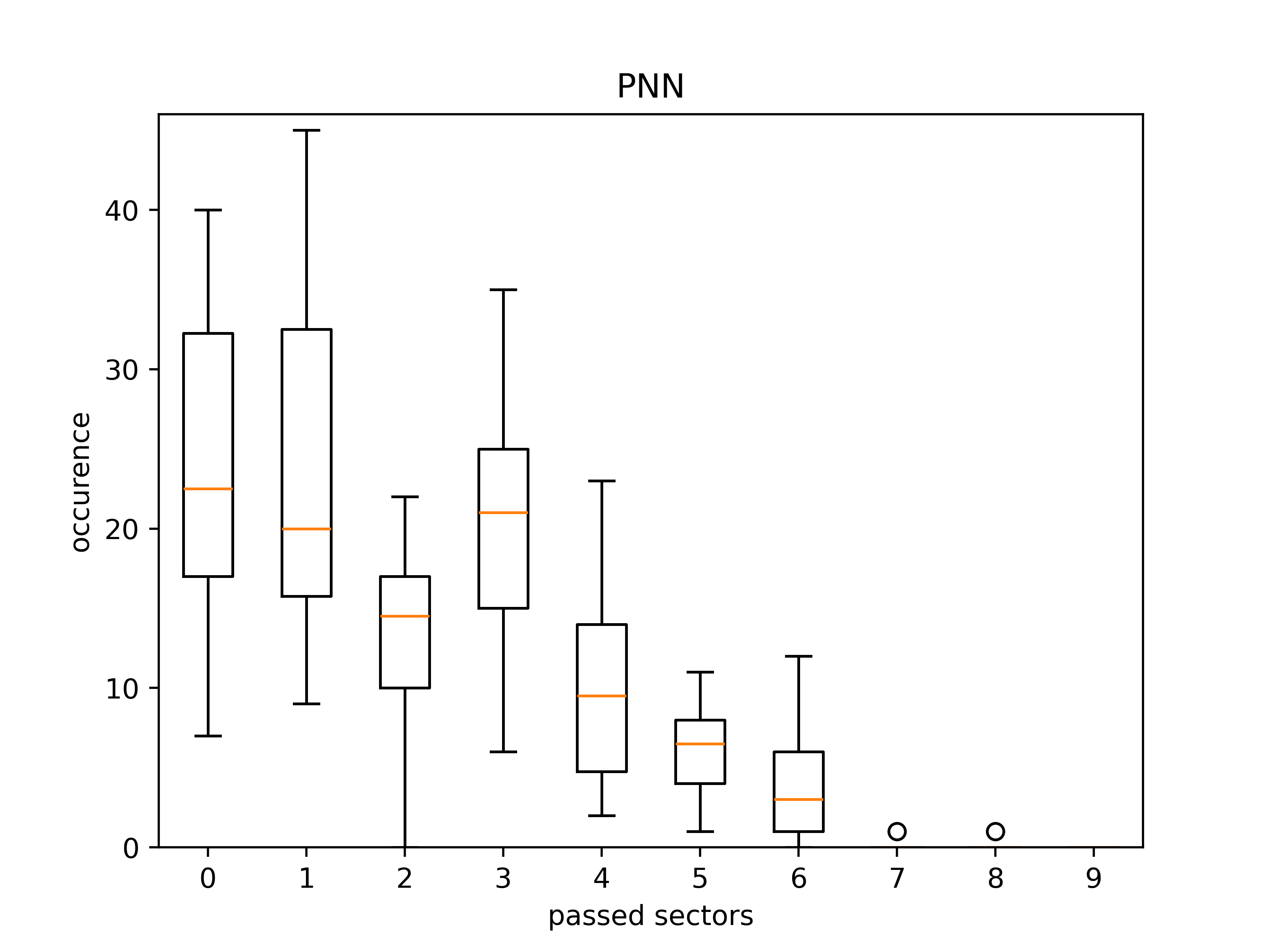}
    }
    \hfill
    \subfloat[]{
        \includegraphics[width=0.4\textwidth]{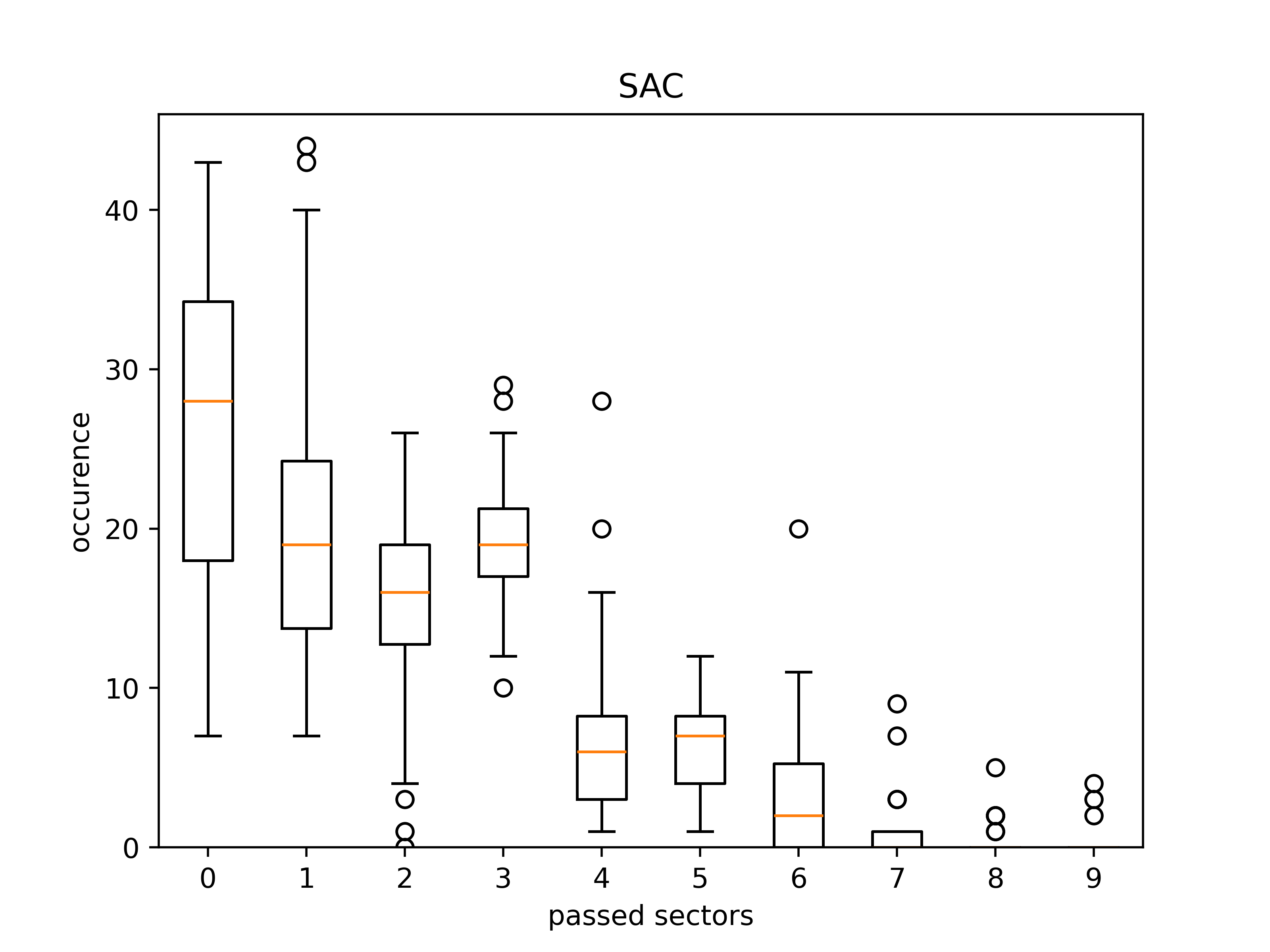}   
    }
    \hfill
    \subfloat[]{
        \includegraphics[width=0.4\textwidth]{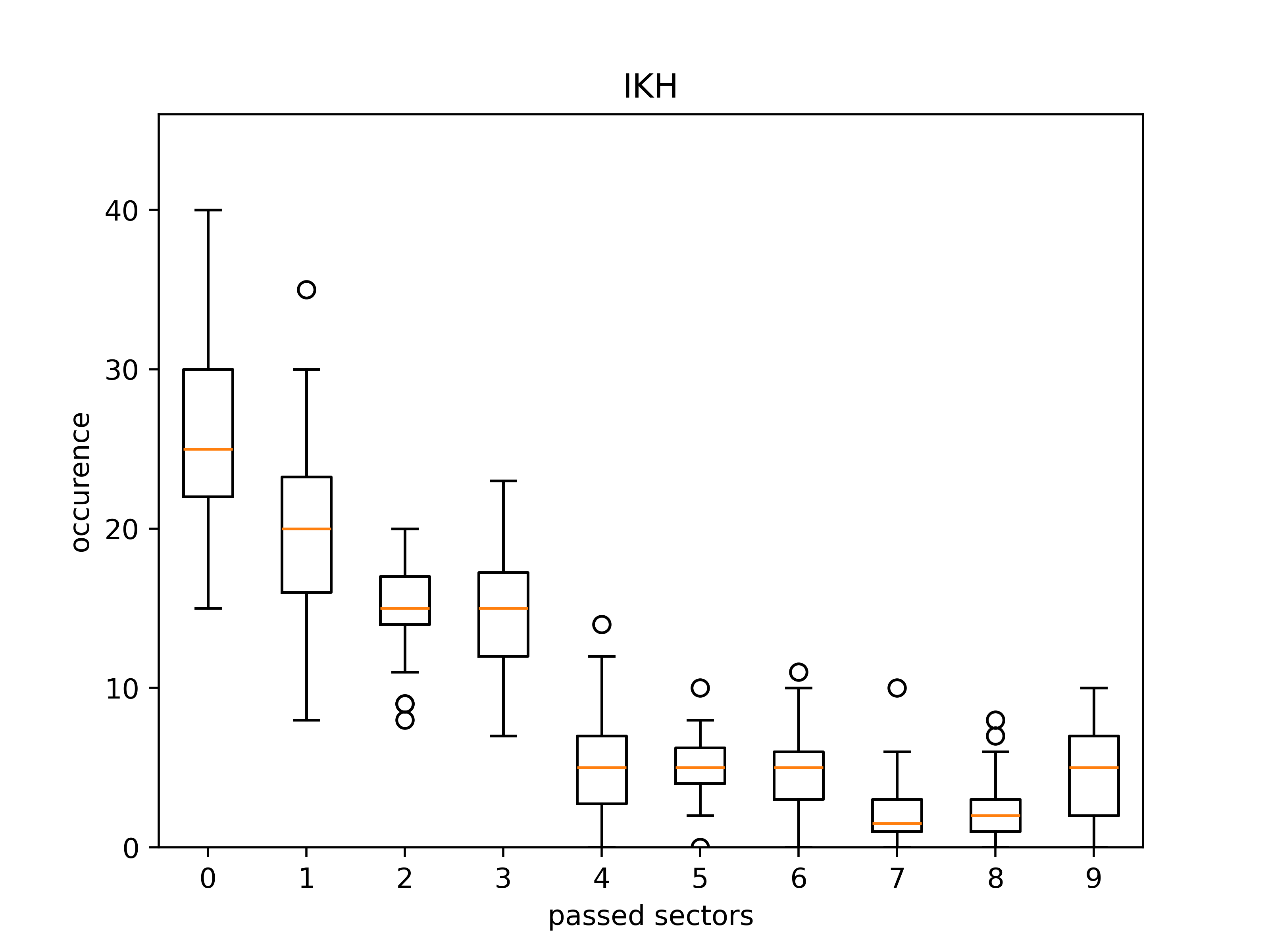}
    }
    \caption{Boxplot indicating the average time of passing $n$ sectors.}
    \label{fig: passed sections}
\end{figure*}

Table \ref{tab: passed sectors statistics} shows briefly the performance of agents completing sectors. In \figurename \ref{fig: passed sections} we report a detailed version of comparisons between algorithms.
The boxplots show the distribution of the number of sectors passed by agents using the indicated algorithm.
The $x$-axis indicates the number of passed sectors, where 9 means 9 or more sectors passed, while the $y$-axis is how many times agents successfully passed the indicated number of sectors.
In general, it is a long-tail distribution, few completed sectors are easier to achieve than completing the entire racetrack.
The performance of agents starts to drop rapidly from completing 4 sectors. Both SAC and PNN hardly ever can complete 7 or more sectors, but IKH can handle it much better and has an increasing occurrence of completing 9 or more sectors.

\begin{figure}[!tb]
    \centering
    \includegraphics[width=0.4\textwidth]{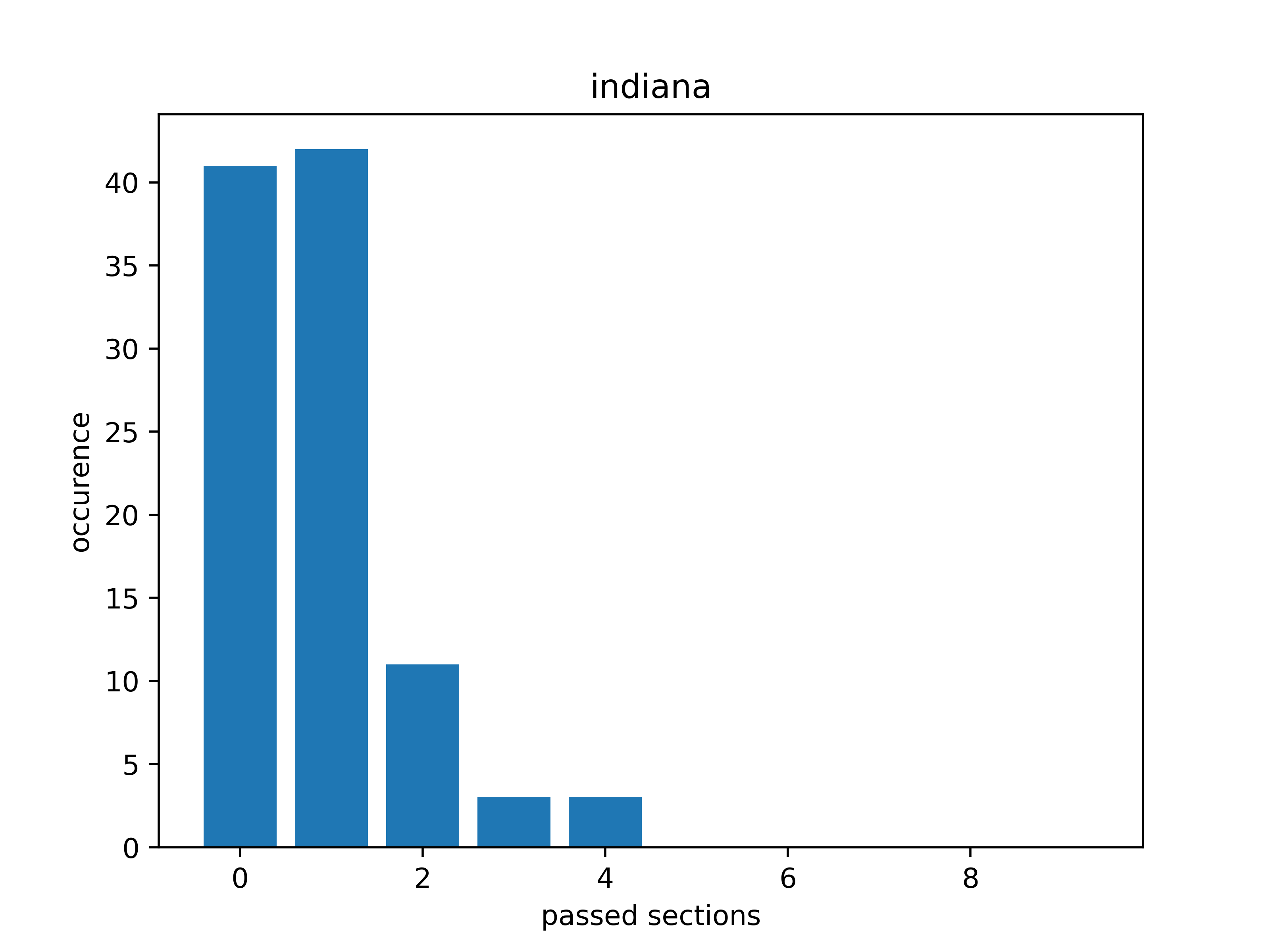}
    \caption{Performance of \texttt{Indiana} on \texttt{Racetrack} scenario}
    \label{fig: indiana on racetrack}
\end{figure}

In \figurename \ref{fig: ablation skills}, different combinations of skills are evaluated to see the performance, and we noticed that well-performing combinations contain the skill \texttt{Indiana}, which is a higher level skill that induces completing a closed racetrack. 
Thus, we evaluated the performance of \texttt{Indiana} on \texttt{Racetrack} scenario. \figurename \ref{fig: indiana on racetrack} shows a barplot of completed sectors, using the same seed of evaluation. 
We can see that \texttt{Indiana} is not able to complete 5 or more sectors, and rarely can achieve 3 and 4 sectors.


\section{On spawning location}
\begin{figure}[!tb]
    \centering
    \includegraphics[width=0.5\textwidth]{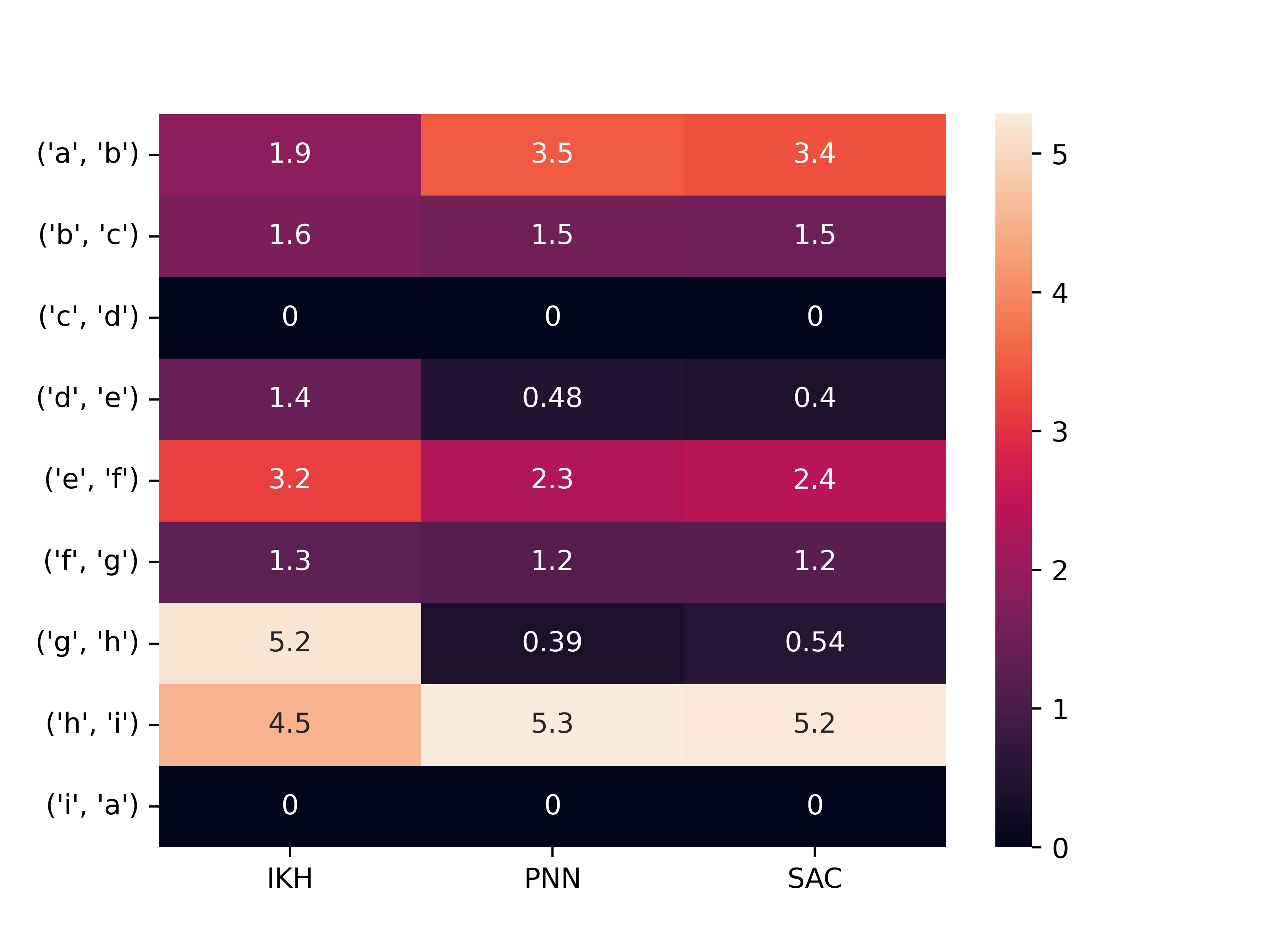}
    \caption{Heatmap representing how the spawn point affects the performance of agents doing the track.}
    \label{fig: heatmap}
\end{figure}

Another factor that affects the performance of the agents is the spawning location. Since the \texttt{Racetrack} is divided into different sectors, we analyzed also what is the performance of each method depending on the spawning location.
We denote each segment in \figurename \ref{fig: racetrack} by a letter starting from $a$. The starting segment road is the long straight road on the bottom, and it is denoted as $(a, b)$, then the next road segment in the counterclockwise sense is $(b, c)$ and so on, and the final segment is $(i, a)$.
In \figurename \ref{fig: heatmap}, the heatmap represents the average number of passed sectors for agents spawned in the indicated sector.
In general, IKH can complete more sectors in almost all situations, except for segments $(a, b)$ and $(h,i)$, where the average is a bit lower. Especially in $(g,h)$ and $(h,i)$ (the big curve on the left side), while SAC and PNN prefer the later road segment, IKH can handle both situations.  
In segments $(c,d)$ and $(i,a)$ the vehicle is never spawned due to the limited length. 

\end{document}